\documentclass[a4paper,fleqn]{cas-dc}

\usepackage[square,sort,comma,numbers]{natbib}
\usepackage{graphicx}
\usepackage{subcaption}
\usepackage{booktabs} 
\usepackage{multirow}
\usepackage{amsmath}
\usepackage{amssymb}
\usepackage{mathtools}
\usepackage{amsthm}
\usepackage{tikz}
\usepackage{pgfplots}
\usepackage{pgfplotstable}
\def\tsc#1{\csdef{#1}{\textsc{\lowercase{#1}}\xspace}}
\tsc{WGM}
\tsc{QE}

\begin{document}
\let\WriteBookmarks\relax
\def\floatpagepagefraction{1}
\def\textpagefraction{.001}

\shorttitle{}    

\shortauthors{}  

\title [mode = title]{FCL-ViT: Task-Aware Attention Tuning for Continual Learning}  



%

\author[1]{Anestis Kaimakamidis}[type=editor,
        orcid=0009-0001-1758-0772
]

\cormark[1]


\ead{akaimak@csd.auth.gr}



\affiliation[1]{organization={Aristotle University of Thessaloniki},
            country={Greece}}

\author[1]{Ioannis Pitas}[type=editor,
        orcid=0009-0006-7555-8641
]

\cormark[1]

\ead{pitas@csd.auth.gr}




\cortext[1]{Corresponding author}

\fntext[1]{}


\begin{abstract}
Continual Learning (CL) involves adapting the prior Deep Neural Network (DNN) knowledge to new tasks, without forgetting the old ones. However, modern CL techniques focus on provisioning memory capabilities to existing DNN models rather than designing new ones that are able to adapt according to the task at hand. This paper presents the novel Feedback Continual Learning Vision Transformer (FCL-ViT) that uses a feedback mechanism to generate real-time dynamic attention features tailored to the current task. The FCL-ViT operates in two Phases. In phase 1, the generic image features are produced and determine where the Transformer should attend on the current image. In phase 2, task-specific image features are generated that leverage dynamic attention. To this end, Tunable self-Attention Blocks (TABs) and Task Specific Blocks (TSBs) are introduced that operate in both phases and are responsible for tuning the TABs attention, respectively. The FCL-ViT surpasses state-of-the-art performance on Continual Learning compared to benchmark methods, while retaining a small number of trainable DNN parameters. 
\end{abstract}


\begin{highlights}
\item The proposal of a novel framework for CL that uses TABs and TSBs as fundamental blocks and, operates in two Phases: 1) Generating generic image features and 2) Generating task-specific features by reprogramming the TABs feature representations dynamically.
\item Extensive experimental FCL-ViT evaluation proves that FCL-ViT goes well beyond the state-of-the-art performance on the image classification benchmarks for CL. 
\item The FCL-ViT performance is compared with other state-of-the-art CL frameworks.
\end{highlights}



\maketitle

\section{Introduction}
Human learning is a challenging concept to understand and analyze. However, the human ability to learn novel tasks, while leveraging the knowledge of previous ones is undeniable. On the other hand, Deep Neural Networks (DNNs) have not yet achieved this summit. As a result, DNN catastrophic forgetting \cite{french1999catastrophic} can occur, when DNNs display a profound decrease in their performance of previous tasks upon learning new ones. This phenomenon can be caused by the feed-forward nature of most DNN architectures involved in CL, which typically cannot provide human ``thinking" options. For example, DNNs cannot reiterate, re-filter, and further analyze a given prompt, while humans reiterate the same thought to come to a safe conclusion. Such a process can be modeled in DNNs using a DNN feedback mechanism. Admittedly, to leverage the knowledge of previous tasks and infer dynamically on incoming images a DNN should be able to assess the nature of the task at hand dynamically. Then the DNN should filter the neural features accordingly to generate on-the-fly an image analysis strategy customized for the incoming image. 

Many recent studies \cite{douillard2020podnet, douillard2022dytox, hou2019learning, rebuffi2017icarl, yan2021dynamically, zhao2020maintaining} have sought to tackle the CL challenges. Among these approaches, some \cite{hou2019learning, rebuffi2017icarl, zhao2020maintaining} have utilized the concept of knowledge distillation \cite{hinton2015distilling}, which involves transferring knowledge from a Teacher DNN model to a Student DNN model to preserve the prior knowledge captured in the network's output logits. In contrast, other methods \cite{douillard2022dytox, yan2021dynamically} have leveraged dynamic expandable networks to address CL challenges. These techniques focus on dynamically enhancing network architectures, such as feature extractors, by incorporating additional DNN parameters and memory. Despite recent progress in tackling the CL problem, forgetting rates reported in CL methods are still high. This occurs because the CL goal is to teach the DNN model where to attend according to the new task at hand. This cannot happen dynamically using the traditional feed-forward DNN architectures. There is a need for  novel DNN architecture that can act sequentially by: 1) assessing the task at hand and, 2) tuning Attention features accordingly. 

To address this challenge, the Feedback Continual Learning Vision Transformer (FCL-ViT) is introduced which operates in two Phases. Phase 1 produces generic image features, i.e. the attention features of the current image. After obtaining the generic image features, phase 2 produces task-specific features to aid the inference of the current image. Thus, FCL-ViT uses Tunable self-Attention Blocks (TABs) that provide the ability to generate both generic and task-specific attention features. FCL-ViT also uses Task Specific Blocks (TSBs) to translate the generic attention features to task-specific attention-tuned features. The TSBs need a CL regularizer,  e.g. Elastic Weight Consolidation (EWC) \cite{kirkpatrick2017overcoming}, to retain previous knowledge while learning a new task. Such a ViT design significantly alleviates catastrophic forgetting and leverages the Attention mechanism feature of tuning the focus according to the needs of the new task. A similar mechanism has worked significantly well for the Transfer Learning (TL) problem, TOAST \cite{shi2023toast} proposed a feedback mechanism that refocuses the Transformer encoder features according to the new task dataset at hand. The key difference with FCL-ViT is that TOAST element-wise adds the value variable of each self-attention layer with the vector produced by the decoder introduced by the TOAST method. On the other hand, the FCL-ViT uses the TSBs to produce vectors that tune features using cross-attention layers. Another difference is that TOAST neither uses a regularizer nor Dropout layers in their implementation, as the nature of their problem (Transfer Learning) is different than CL.

The key contribution of this paper is the proposal of a novel framework for CL that uses TABs and TSBs as fundamental blocks and, operates in two Phases: 1) Generating generic image features and 2) Generating task-specific features by reprogramming the TABs feature representations dynamically. Extensive experimental FCL-ViT evaluation proves that FCL-ViT goes well beyond the state-of-the-art performance on the image classification benchmarks for CL. More specifically, extensive experiments are conducted on different splits of the widely used image classification benchmarks Imagenet-100 and CIFAR100 to display the benefits of FCL-ViT in a CL setting. The FCL-ViT performance is compared with other state-of-the-art CL frameworks such as iCaRL \cite{rebuffi2017icarl}, WA \cite{zhao2020maintaining}, DER \cite{yan2021dynamically}, and DyTox \cite{douillard2022dytox}.

The rest of this paper is structured in the following way. Section \ref{sec:relatedWork} presents the state-of-the-art methods on CL, primarily focused on methods using Transformers. Section \ref{sec: methodology} describes the architecture of FCL-ViT and the setup of the CL problem this method is tackling. Section \ref{sec: experiments} exhibits the state-of-the-art performance of FCL-ViT on the benchmark datasets Imagenet and CIFAR100 and ablates on the influence of the TSBs on their impact on catastrophic forgetting. Moreover, the Section includes ablation studies on the relative influence of the hyperparameter $\lambda$ on its ability to retain previously learned knowledge.

\section{Related Work}
\label{sec:relatedWork}

Continual Learning methods can be summarized into three main categories: regularization-based techniques, replay-based techniques, and dynamic expandable networks.

Regularization-based methods primarily use a regularizer to mitigate vast changes to the most important weights per task. A regularization-based method is Elastic Weight Consolidation (EWC), which calculates important weights and adds a regularizer loss to prevent these weights from changing upon learning new tasks. Another approach is called Learning without Forgetting (LwF) \cite{li2017learning} where the DNN model is jointly optimized both for obtaining high classification accuracy in the new task and for preserving accuracy in the old one, without requiring access to the old training data. This method actually employs KD to achieve CL, since it involves optimizing the model on new data according to both ground truth and to the original network's response to the new data.

In contrast to previously described CL methods, replay-based CL approaches retain a portion of previous data, referred to as \emph{exemplars}. Subsequently, a DNN model is trained on both a new training dataset and the stored exemplars to prevent catastrophic forgetting of previous tasks. One such method is Incremental Classifier and Representation Learning (iCaRL) \cite{rebuffi2017icarl}, which utilizes a memory buffer derived from LwF \cite{li2017learning}. iCaRL employs a herding data sampling strategy in classification, where mean data features for each class are calculated, and exemplars are iteratively selected to bring their mean closer to the class mean in feature space. An alternative algorithm is End-to-End Incremental Learning (EEIL) \cite{castro2018end}, which introduces a balanced training stage to finetune the DNN model on a dataset with an equal number of exemplars from each class encountered thus far.

Dynamic expandable networks, such as DER \cite{yan2021dynamically}, offer another strategy by dynamically expanding the model's architecture with a new feature extractor for each task, which then feeds into a unified classifier. Although this method has achieved state-of-the-art performance, the continuous expansion of the model can lead to excessive overhead, and the pruning methods used to manage this, like HAT \cite{serra2018overcoming}, demand high computational resources. Furthermore, while DER eliminates the need for task identifiers during inference, its hyperparameter-sensitive pruning poses practical challenges in real-world applications. In contrast, more recent works, such as Simple-DER \cite{li2021preserving}, aim to resolve these limitations by adopting dynamic architectures specialized for each task. Though effective, they either struggle with uncontrolled parameter growth or require hyperparameter tuning.

While vision transformers have achieved notable success, their use in Continual Learning remains relatively unexplored. Recent methods like DyTox \cite{douillard2022dytox} aim to address this by expanding task-specific tokens or employing inter-task attention mechanisms to incorporate previous task information, preventing performance degradation between tasks. However, these approaches require additional memory for storing past tasks' training instances. Moreover, many approaches, such as L2P \cite{wang2022learning} and DualPrompt \cite{wang2022dualprompt}, rely on soft prompts and thus present scalability issues when the number of tasks increases.

Transformers were first introduced for machine translation tasks \cite{vaswani2017attention} Since then, they have become the state-of-the-art model for various Natural Language Processing (NLP) tasks \cite{devlin2018bert, girdhar2019video}. The key element of the Transformer is the attention module, which gathers information from the entire input sequence. Recently, the Vision Transformer (ViT) \cite{dosovitskiy2020image} adapted the Transformer architecture for image classification, achieving scalability when working with large datasets. Following this, significant efforts have been made to enhance Vision Transformers in terms of data and model efficiency \cite{han2020survey, khan2022transformers, yuan2021tokens}. A prominent research direction involves incorporating explicit convolutions or convolutional properties into the Transformer architecture \cite{dai2021coatnet, vaswani2021scaling}. For example, CoaT \cite{xu2021co} introduces a conv-attention module that uses convolutions to implement relative position embeddings. LeViT \cite{graham2021levit} adopts a pyramid structure with pooling to capture convolutional-like features instead of the uniform Transformer architecture. Similarly, CCT \cite{hassani2021escaping} removes the need for class tokens and positional embeddings by using a sequence pooling strategy and convolutions instead.

\section{FCL-ViT Methodology}

\label{sec: methodology}

\begin{figure*}[h]
    \centering
    \includegraphics[width=0.8\textwidth]{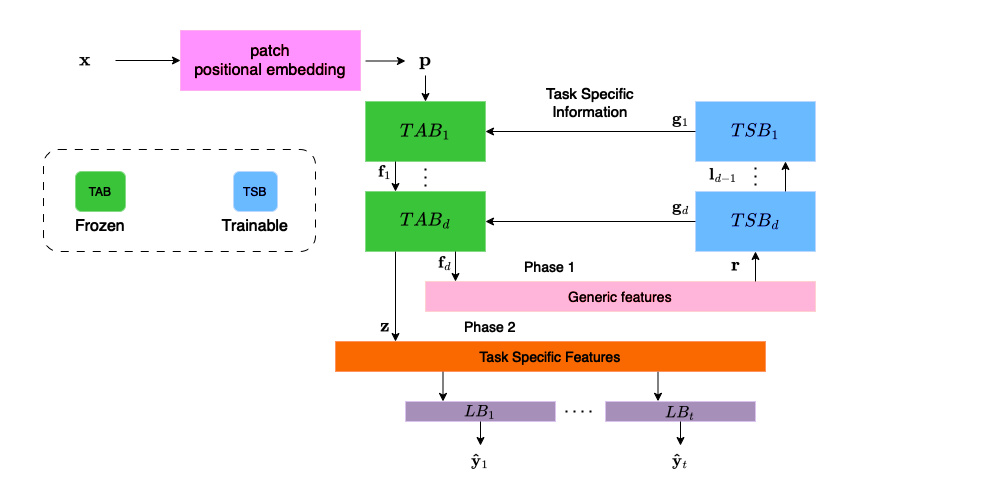}
    \caption{FCL-ViT architecture.}
    \label{fcl-vit}
\end{figure*}

\subsection{Problem Setup}
 The CL goal is to progressively update a single DNN image classification model to accommodate new classes as they become available. In a series of classification tasks $T_1, T_2, \dots, T_t$, task dataset $\mathcal{D}_i = \left\{(\mathbf{x}_j^i, y_j^i)\right\}_{j=1}^N$ comprises $N_i$ samples, and $\mathbf{x}_j^i$ and $y_j^i$ denote the $j$-th image and its corresponding label. One task may comprise data coming from various classes. The class set associated with the training task $T_j$ is denoted by $\mathcal{C}_j$. We assume that there is no overlap between the classes of different tasks, meaning that $\forall i, j, \mathcal{C}_i \cap \mathcal{C}_j = \emptyset$. For the current task, only samples from $\mathcal{C}_t$, are available as this method is not using any kind of rehearsal memory. The DNN model is evaluated on a union of test datasets $\mathcal{Z} = \mathcal{Z}_1 \cup \dots \cup \mathcal{Z}_t$, and it is expected to classify all the classes belonging to the overall class set $\mathcal{C}_{1\sim t} = \mathcal{C}_1 \cup \dots \cup \mathcal{C}_t$, but using the task index $i$.

\subsection{Feedback Continual Learning Vision Transformer (FCL-ViT)}

Most DNN models are feed-forward ones. Therefore, DNN inference is performed entirely on one forward pass. As CL aims to reduce catastrophic forgetting upon learning new tasks, this is a rather hard task for feed-forward classification DNNs, as no feedback is provided upon inference. To address CL needs, this paper proposes a novel CL ViT architecture, the Feedback Continual Learning Vision Transformer (FCL-ViT). The FCL-ViT is a Transformer architecture using feedback for CL classification problems. The FCL-ViT architecture, shown in Figure \ref{fcl-vit}, it comprises Tunable self-Attention Blocks (TABs) and Task-Specific Blocks (TSBs) for image feature extraction. TSBs are entirely responsible for dynamically tuning the FCL-ViT features during inference. 

The FCL-ViT architecture uses a feedback mechanism to tune the features during inference. FCL-ViT Inference is conducted in two phases: a) the generic image features phase and b) the task-specific features phase. As a result, given a data sample $\mathbf{x}$, the patch positional embeddings $\mathbf{p} \in \mathbf{R}^{D}$ are produced, where $D$ denotes the dimension. In FCL-ViT Phase 1, $\mathbf{p}$ passes through $d$ TABs to generate the generic image features representations $\mathbf{r}$. The representations $\mathbf{r}$ are passed through the TSBs to generate the task-specific features for TABs $\left\{\mathbf{g}_j, j = 1, ..., d\right\}$, where $d$ denotes the ViT depth. In Phase 2, $\mathbf{p}$ is passed again through the TABs to generate the task-specific features $\mathbf{z}$, to be used for classifying image $\mathbf{x}$. For simplicity, the following equations describe the architecture using one attention head per block.

The TABs play a dual role as they participate in both FCL-ViT phases. The structure of a TAB allows it to switch between the two phases. As a result, a TAB is a typical self-attention module in Phase 1:
\begin{equation}
\begin{split}
& \mathbf{Q}_j = \mathbf{W}_{qj} \mathbf{f}_{j-1},  \\
& \mathbf{K}_j = \mathbf{W}_{kj} \mathbf{f}_{j-1},   \\
& \mathbf{V}_j = \mathbf{W}_{vj} \mathbf{f}_{j-1},  \\
& \mathbf{A}_j = Softmax \left( \frac{\mathbf{Q}_j \mathbf{K}_j^T}{\sqrt{D/h}} \right),  \\
& \mathbf{f}_{j} = \mathbf{W}_{oj} \mathbf{A}_j \mathbf{V}_j + b_o, \mathbf{f}_o = \mathbf{p}, \mathbf{r} =\mathbf{f}_d \in \mathbf{R}^{D}.
\end{split}   
\end{equation}

Where, $f_j, j = 1, ..., d$ are the TABs outputs during phase 1. The generic image vector $\mathbf{r}$ is equal to the $d$-th TAB output  $\mathbf{r} =\mathbf{f}_d$. During, phase 2 the $j$-th TAB operates as a typical cross-attention module between $f_{j-1}$ and the output $g_j$ of the $j$-th TSB: 

\begin{equation}
\begin{split}
& \mathbf{Q}_j = \mathbf{W}_{qj} \mathbf{f}_{j-1}, \\
& \mathbf{K}_j = \mathbf{W}_{kj} \mathbf{g}_j, \\
& \mathbf{V}_j = \mathbf{W}_{vj} \mathbf{g}_j, \\
& \mathbf{A}_j = Softmax \left( \frac{\mathbf{Q}_j \mathbf{K}_j^T}{\sqrt{D/h}} \right), \\
& \mathbf{f}_{j} = \mathbf{W}_{oj} \mathbf{A}_j \mathbf{V}_j + b_o, \mathbf{f}_0 = \mathbf{p}, \mathbf{z} = \mathbf{f}_d \in \mathbf{R}^{D}.
\end{split}
\end{equation}

At the end of the second phase, the finetuned feature vector $\mathbf{z}$ of $\mathbf{x}$ is produced for image classification. The TSBs have a specific structure to capture the task-specific information and refine the TAB features. The $j$-th TSB consists of two linear layers each equipped with a Dropout function. The first one filters the output of the previous TSB $\mathbf{l}_j$ to be passed to the TAB in the same depth $j = 1,...,d$, and the other one produces the output $\mathbf{l}_{j-1}$ to be passed to the next TSB $j-1$:

\begin{equation}
\begin{split}
& \mathbf{g}_j = Dropout (\mathbf{H}_{1j} \mathbf{l}_j; p), \\
& \mathbf{l}_{j-1} = Dropout (\mathbf{H}_{2j} \mathbf{l}_j; p), \mathbf{l}_d = \mathbf{r} \in \mathbf{R}^{D}\\
\end{split}
\end{equation}

Where $p$ is the dropout rate. 

When a new task is learned a new Linear Block (LB) is added to classify image $\mathbf{x}$ using the finetuned feature vector $\mathbf{z}$. The decision is taken using the respective LB for the task $\hat{\mathbf{y}}_k = \Tilde{\mathbf{H}}_k \mathbf{z}, k \leq i$, where, $\Tilde{\mathbf{H}}_k \in \mathbf{R}^{DxC_i}$. The trainable parameters $\theta_j, j=1,...,d$ denotes all entries of the $\mathbf{H}_{1j}$ and $\mathbf{H}_{2j}$ matrices. The parameters $\Tilde{\mathbf{H}}_k$ are also trainable, but are trained only during the $k$-th task.

Using this structure, the TABs can initialize their weights using a publicly available dataset e.g. the ImageNet \cite{deng2009imagenet}, operating in Phase 1. The TABs weights remain frozen and are used as a backbone. This ensures that the TABs produce general image features in Phase 1. Then, the architecture is ready to learn tasks continually without forgetting older tasks by freezing the TAB weights and only training the TSB modules. When we retain the $j$-th TSB to learn a new task $i$, $i = 1,...,t$, the respective parameter vector is denoted by $\theta_{i,j}$. To ensure that the TSBs retain the previous task information a regularizer is needed, e.g. the Elastic Weight Consolidation (EWC) \cite{kirkpatrick2017overcoming}, which assigns an update penalization factor per TSB parameter in $\mathbf{H}_{1j}$, $\mathbf{H}_{2j}$, based on their importance for previously learned tasks, as quantified using the Fisher Information Matrix $\mathbf{F}$: 
\begin{equation}
    \mathcal{L}_c = \sum_j \sum_l \frac{\lambda}{2}F_{ll}(\theta_{i,j,l} - \theta_{i-1, j,l})^2,
\label{loss_cont}
\end{equation}
\noindent 
where $\theta_{i,j}$ and $\theta_{i-1,j}$ denote the TSB trainable parameters $\theta_{j}, j=1,...,d$  when the DNN is trained for tasks $T_i$ and $T_{i+1}$ respectively. The hyperparameter $\lambda$ sets the importance of previous tasks compared to the new ones and $l$ indexes each DNN parameter. Any change to the $l$-th DNN parameter is penalized by a factor proportionate to the $l$-th diagonal entry $F_{ll}$ of $\mathbf{F}$, evaluated once on the parameter set upon which training on the old task originally converged. The method exploits the fact that $\mathbf{F}$ can be computed from first-order derivatives alone while assuming that a multivariate Gaussian probability distribution approximation to the posterior distribution $p(\theta_i|\mathcal{D}_i)$ of the model parameters conditioned on the old task's training dataset $\mathcal{D}_i$ suffices.

The FCL-ViT training procedure is conducted using all available samples per task sequentially and without using any form of rehearsal memory or replaying samples from previous tasks. The FCL-ViT generates its prediction for a sample $\mathbf{x}$ and this prediction is fed into the loss function: 

\begin{equation}
    \mathcal{L} = \mathcal{L}_c  + \alpha \mathcal{L}_h(\mathbf{y}, \hat{\mathbf{y}}), 
\end{equation}

\noindent where $\hat{\mathbf{y}}$ is the FCL-ViT model prediction for input $\mathbf{x}$, $\mathbf{y}$ is the respective one-hot-encoded ground-truth label, $\mathcal{L}_h(\cdot,\cdot)$ is the cross-entropy loss, and $\alpha$ is a hyperparameter controlling the relative influence of the main cross-entropy loss term. 

\section{Experimental FCL-ViT Evaluation}

\label{sec: experiments}

\subsection{Comparison with state-of-the-art methods}

Following common CL evaluation protocols \cite{rebuffi2017icarl, douillard2022dytox, yan2021dynamically}, the FCL-ViT is evaluated on different splits of the CIFAR-100 \cite{krizhevsky2009learning}. A pretrained ViT \cite{dosovitskiy2020image} base is employed as a backbone to initialize the TABs weights. The chosen FCL-ViT contains $d=12$ TABs with an embedding dimension of $D=768$ and 12 attention heads. The FCL-ViT also contains $d=12$ TSBs with a dropout rate of $p=0.5$. After a hyperparameter search, the batch size was set to 128 and Adam was adopted as an optimizer, with an initial learning rate of $10^{-3}$. The number of epochs was set to 100. Different performances are measured for Imagenet-100 on 10 tasks (10 new classes per task). Table \ref{tab:comparison_table_imagenet} presents the average of the Top-1 classification accuracies (``Avg") after every task as defined in \cite{rebuffi2017icarl} and also the Top-1 classification accuracy after the last task is reported (``Last"). ``\#TP" denotes the number of trainable parameters for each method in millions. The same performances are also measured for the CIFAR100 dataset on 10 tasks (10 new classes per task), 20 tasks (5 new classes per task), and 50 tasks (2 new classes per task). Table \ref{tab:comparison_table} presents the relevant results, comparing FCL-ViT to other CL approaches. Benchmark results contained in Table \ref{tab:comparison_table_imagenet} and Table \ref{tab:comparison_table} come from \cite{douillard2022dytox}.

\begin{table}
    \centering
    \caption{Top-1 accuracy classification results on Imagenet-100 for 10 task splits.}
    \label{tab:comparison_table_imagenet}
    
    \begin{tabular}{lccccccccc}
        \toprule
        & \multicolumn{3}{c}{10 tasks}  \\
        \cmidrule(lr){2-4} 
        Methods & \#TP & Avg & Last  \\
        \midrule

        iCaRL \cite{rebuffi2017icarl} & 11.22 & 57.84 &  46.13 \\
        WA \cite{zhao2020maintaining} & 11.22 & 69.44  & 53.78 \\
        DER \cite{yan2021dynamically} & 112.27 & \textbf{77.18}  & 66.70 \\
        DyTox+ \cite{douillard2022dytox} & 11.01 & 77.15 & 69.10  \\
        FCL-ViT & 14.23 & 75.42 & \textbf{71.80}  \\
        
        \bottomrule
    \end{tabular}
    
\end{table}

\begin{table*}[htbp]
    \centering
    \caption{Top-1 accuracy classification results on CIFAR100 for task splits.}
    \label{tab:comparison_table}
    \resizebox{\textwidth}{!}{%
    \begin{tabular}{lccccccccc}
        \toprule
        & \multicolumn{3}{c}{10 tasks} & \multicolumn{3}{c}{20 tasks} & \multicolumn{3}{c}{50 tasks} \\
        \cmidrule(lr){2-4} \cmidrule(lr){5-7} \cmidrule(lr){8-10}
        Methods & \#TP & Avg & Last & \#TP & Avg & Last & \#TP & Avg & Last \\
        \midrule

        iCaRL \cite{rebuffi2017icarl} & 11.22 & 65.27 $\pm$ 1.19 & 50.74 & 11.22 & 61.00 $\pm$ 2.03 & 43.75 & 11.22 & 56.08 $\pm$ 3.08 & 36.62 \\
        WA \cite{zhao2020maintaining} & 11.22 & 69.44 $\pm$ 1.29 & 53.78 & 11.22 & 67.33 $\pm$ 1.13 & 47.31 & 11.22 & 62.08 $\pm$ 1.08 & 42.14 \\
        DER \cite{yan2021dynamically} & 112.27 & 75.36 $\pm$ 0.65 & \textbf{65.22} & 224.55 & 74.09 $\pm$ 0.84 & 62.48 & 561.39 & 72.41 $\pm$ 0.90 & 59.08 \\
        DyTox+ \cite{douillard2022dytox} & 10.73 & 75.54 $\pm$ 1.04 & 62.06 $\pm$ 0.14 & 10.74 & 75.04 $\pm$ 0.88 & 60.03 $\pm$ 1.06 & 10.77 & 74.35 $\pm$ 1.45 & 57.09 $\pm$ 0.26 \\
        FCL-ViT & 14.23 & \textbf{77.63} $\pm$ 0.42 & 65.02 $\pm$ 0.05 & 14.23 & \textbf{76.60} $\pm$ 0.31 & \textbf{67.61} $\pm$ 0.12 & 14.23 & \textbf{ 78.59} $\pm$ 0.56 & \textbf{67.72} $\pm$ 0.15 \\
        
        \bottomrule
    \end{tabular}
    }
\end{table*}

\begin{figure*}[ht]
    \centering
    
    \includegraphics[width=\textwidth]{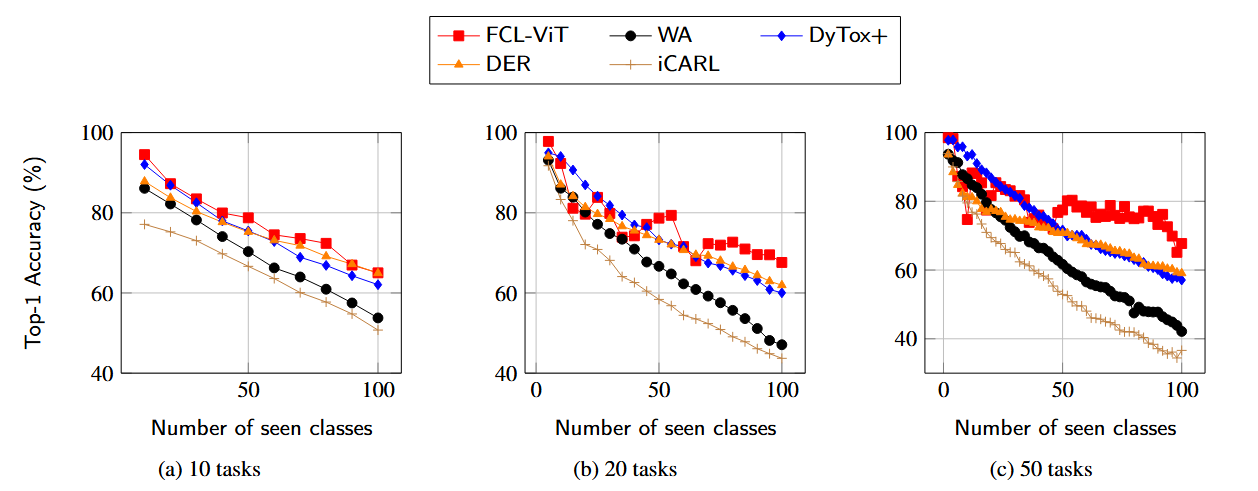}

    \caption{The Top-1 classification accuracy (\%) is reported after learning each task. FCL-ViT outperforms other models for all CIFAR100 splits.}
    \label{fig:diff_steps}
\end{figure*}

On the Imagenet-100 dataset the FCL-ViT outperforms all competing methods on the ``Last" classification accuracy, which is the desired result from a CL method, as it measures the final average accuracy of all tasks. However, it presents a bit lower ``Avg" classification accuracy than two of the competing methods. This means that these methods leveraged better the training dataset throughout the training  process and had higher performance while learning the tasks, but the  ``Last" classification accuracy dropped significantly which means that they also displayed higher forgetting rates. Regarding the CIFAR100 dataset, FCL-ViT outperforms all competing methods, except for the ``Last" classification accuracy for the 10 task problem. The difference between the DER classification accuracy is 0.2\% while DER uses x7.89 more trainable parameters than the FCL-ViT. Although all other methods' performance is decreased with increasing task numbers on both the ``Avg" and the ``Last" classification accuracy, FCL-ViT performance is stable, around $77\%$ and $67\%$ for ``Avg" and ``Last" classification accuracy, respectively. This proves that the FCL-ViT architecture is tolerant to learning multiple tasks continually without exhibiting catastrophic forgetting behavior. The same results are seen in Figure \ref{fig:diff_steps} where the average Top-1 accuracy after learning a task is reported for FCL-ViT and all benchmark methods. As the number of seen classes increases, FCL-ViT retains its classification accuracy much better than its competitors. It is noted here that most of the benchmark methods use rehearsal memory to avoid catastrophic forgetting. FCL-ViT replays none of the samples learned in previous tasks.

FCL-ViT achieves superior performance compared to the baseline methods because of the nature of regularization methods and the Transformer adaptability. The performance boost comes from the fact that the EWC method operates in a higher-dimensional space, allowing it to better discriminate FCL-ViT features across tasks. This level of discrimination is not possible in traditional feed-forward DNNs, which are limited by the lower dimensionality of their latent space. Additionally, the ability to tune a Transformer's features and steer the Attention allows the FCL-ViT to perform this better than its competitors.

\begin{figure}[!ht]
    \centering
    \includegraphics[width=0.45\textwidth]{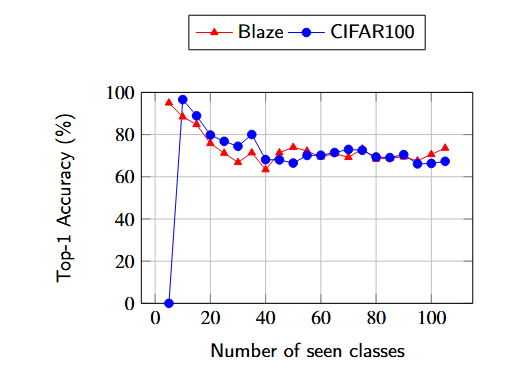}
    \caption{FCL-ViT Top-1 classification accuracy on BLAZE (5 classes) and CIFAR100 (100 classes) datasets. The FCL-ViT learns BLAZE dataset first and then CIFAR100 with 20 tasks (5 classes for each task).}
    \label{fig:blaze}
\end{figure}

To showcase the extended capabilities of the FCL-ViT architecture, it is showcased in a Natural Disaster Management (NDM) scenario. The BLAZE wildfire classification dataset is employed, which contains wildfire pictures. Such images have significant differences in the input images from CIFAR100 image entries. The BLAZE classification dataset contains 3850 train images and 1567 test images for the classification of areas during or after a wildfire. The classes are Fire, Burnt, Half-Burnt, Non-Burnt and Smoke. As a result, the experiment using CIFAR100 with 20 tasks (5 new classes per task) is repeated after FCL-ViT has first learned the BLAZE classification dataset. Figure \ref{fig:blaze} displays the Average Top-1 Accuracy of the BLAZE and the CIFAR100 datasets respectively as the FCL-ViT learns the new tasks. The conclusions of this experiment are two-fold: a) the FCL-ViT did not fall for catastrophic forgetting of the BLAZE dataset after learning 100 classes of the CIFAR100 dataset, and b) the CIFAR100 dataset displayed a small reduction from the average Top-1 Accuracy from 67.61 \% to 67.32 \% without and after learning the BLAZE wildfire dataset first respectively, measured after learning the last task.

\subsection{Ablation Studies}
\begin{figure}[!ht]
    \centering
    \includegraphics[width=0.45\textwidth]{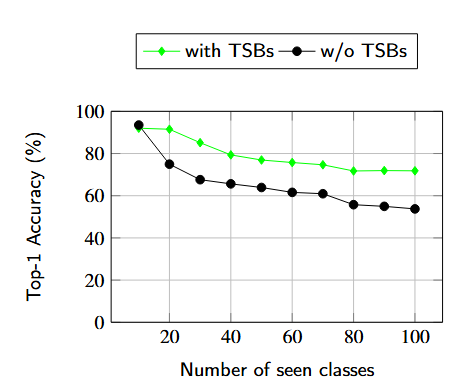}
    \caption{FCL-ViT classification accuracy on the Imagenet-100 dataset with and without the TSBs.}
    \label{fig:nofeed}
\end{figure}

The main contribution of this method is that it showcases how a feedback mechanism can be used to filter the general transformer features into task-specific features, while avoiding catastrophic forgetting. The component responsible for this mechanism is the TSB, which transforms the general features into vector inputs for the main transformer to produce task-specific features. To this end, we measured the impact of the TSBs on the CL scenario of learning the Imagenet-100 in 10 tasks (10 new classes per task). The TSBs are removed and the ViT base is only trained using the EWC regularizer on all of the weights. The results are presented in Figure \ref{fig:nofeed}. As expected the FCL-ViT without the TSBs is more susceptible to forgetting as higher forgetting rates are demonstrated in this experiment. The ``Last" accuracy of the FCL-ViT is 71.80\% and 53.75\% with the TSBs and without the TSBs respectively. The influence of the TSBs in mitigating forgetting is high because of their ability to reprogram the features and produce task-specific features during the second phase.

\begin{figure}[!ht]
    \centering
    \includegraphics[width=0.45\textwidth]{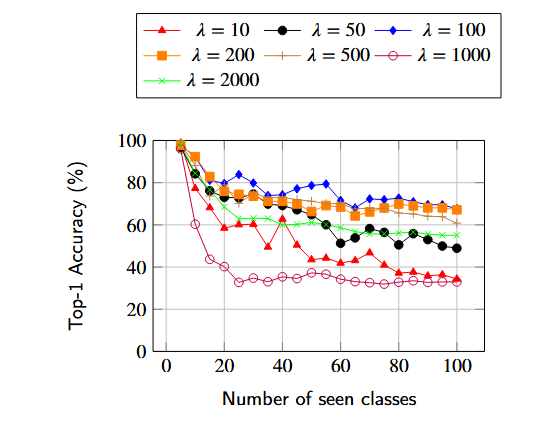}
    \caption{Effect of hyperparameter $\lambda$ on FCL-ViT classification accuracy}
    \label{fig:ablation}
\end{figure}

The EWC regularizer plays a crucial part in retaining TSB previous knowledge. To study this further, the $\lambda$ hyperparameter is ablated. This experiment examines how strong the EWC regularizer influence should be on TSB training. The ablation experiment is the same as the one presented in Figure \ref{fig:diff_steps} with the 20 tasks, for various $\lambda$ hyperparameter values. The ablation study results are shown in Figure \ref{fig:ablation} and are quite expected. The small $\lambda$ values cause the first tasks learned to be forgotten. Thus the average Top-1 Accuracy decreases. On the other hand, the highest $\lambda$ values constrain the gradients during learning. Therefore, FCL-ViT fails to achieve high classification performance on new tasks, while retaining knowledge of previous tasks. The best FCL-ViT classification performances were attained with intermediate $\lambda$ values ( $\lambda = 100$ and $\lambda = 200$) that balance the stability-plasticity FCL-ViT trade-off.

\subsection{Why FCL-ViT works better?}
The FCL-ViT introduces a two-phase approach that tunes the general transformer features into task-specific features, and achieves state-of-the-art Continual Learning performance. When combined with the EWC method, the FCL-ViT can retain the feature tuning needed to maximize performance for each task. Compared to distillation-based methods \cite{rebuffi2017icarl, li2017learning} the FCL-ViT learns the feature mapping that helps each task maximize the respective accuracy, while these methods replay model logits and data from previous tasks. HAT and DER methods \cite{serra2018overcoming, yan2021dynamically} dynamically add new feature extractors for each task, which is close to the FCL-ViT approach. Finally, Dytox+ \cite{douillard2022dytox} stores the task-specific information on tokens which can be used upon inference. FCL-ViT performs better than the first category because the optimization space is expanded, i.e. more parameters are tuned for each specific task, while the general features are frozen and tuned using task-specific information. Replay-based and distillation methods constantly change the feature space with a mix of old and new task features and this is the reason they demonstrate high forgetting rates in large sets of tasks. The FCL-ViT outperforms the rest of the methods because of the cross-attention addition and the ability to dynamically adapt to each task. These methods are close to the FCL-ViT nature, but they do not use knowledge from previous task features or tokens to help them on the current task and this is the main reason why FCL-ViT outperforms them. Another issue with these methods is that as the number of tasks is increased their trainable parameters need to be increased, which is not applied to the FCL-ViT. On the other hand, the two-phase approach introduces complexity during inference. The inference of the FCL-ViT is approximately 92 \% slower than the inference of a ViT base which is the cost of the performance gain and the dynamic feature adaptation this method introduces. 

\subsection{Training Stability}
The training stability of the model is ensured by the design of the FCL-ViT. The TSBs and the two-phase approach do not introduce training instabilities due to the following reasons. Firstly, only the TSBs are trainable while the TABs are frozen. The CL problem is now translated into task-specific feature tuning and thus only the simple TSB structure is trained. Secondly, the use of the EWC regularizer constrains the weights and ensures stability. The  Fisher Information Matrix $\mathbf{F}$ penalizes any big changes to important weights ($\theta_{i-1, j,l}$) from the last task $i-1$. As a result, instead of updating the weight values arbitrarily, a selective penalization is applied to ensure training stability. Last but not least, the use of the dropout layers in the TSB design further ensures training stability and further enables the EWC method to retain previous knowledge.

\section{Conclusion}

In conclusion, this work introduces FCL-ViT, a robust Vision Transformer-based framework that addresses the unique challenges of Continual Learning (CL) through dynamic task-specific attention mechanisms. By employing Tunable self-Attention Blocks (TABs) and Task Specific Blocks (TSBs), FCL-ViT effectively adapts to new tasks while preserving previously learned information, thereby reducing catastrophic forgetting. The dual-phase structure leverages both general and task-specific attention, which allows for targeted and efficient learning across diverse datasets, as demonstrated through extensive evaluations on Imagenet-100, CIFAR100, and BLAZE datasets.

The results consistently highlight FCL-ViT’s superior performance in comparison to established CL methods, showcasing its resilience and adaptability across multiple task splits. This resilience underscores the effectiveness of the feedback mechanism and EWC-based regularization in maintaining prior knowledge. Moreover, the model’s capability to operate without rehearsal memory marks a significant advancement in CL, particularly in real-world scenarios, e.g. in embedded computing, where memory constraints are critical. 

\section{Acknowledgement}
The research leading to these results has received funding from the European Commission - European Union (under HORIZON EUROPE (HORIZON Research and Innovation Actions) under grant agreement 101093003 (TEMA) HORIZON-CL4-2022-DATA-01-01). Views and opinions expressed are however those of the authors only and do not necessarily reflect those of the European Union - European Commission. Neither the European Commission nor the European Union can be held responsible for them.




\bibliographystyle{cas-model2-names}

\bibliography{example_paper}



\end{document}